\documentclass[runningheads]{llncs}

\usepackage[T1]{fontenc}
\usepackage{graphicx}
\usepackage{amsmath}
\usepackage{amssymb}
\usepackage{booktabs}  
\usepackage{url}
\usepackage{hyperref}

\begin{document}


\title{IntSeqBERT: Learning Arithmetic Structure in OEIS
       via Modulo-Spectrum Embeddings}

\titlerunning{IntSeqBERT: Learning Arithmetic Structure in OEIS}

\author{Kazuhisa Nakasho\inst{1}}

\authorrunning{K. Nakasho}

\institute{Iwate Prefectural University, Takizawa, Iwate, Japan\\
\email{nakasho\_k@iwate-pu.ac.jp}}

\maketitle


\begin{abstract}
Integer sequences in the OEIS span values from single-digit constants to astronomical
factorials and exponentials, making prediction challenging for standard tokenised models that
cannot handle out-of-vocabulary values or exploit periodic arithmetic structure.
We present \textbf{IntSeqBERT}, a dual-stream Transformer encoder for masked integer-sequence
modelling on OEIS. Each sequence element is encoded along two complementary axes: a continuous
log-scale magnitude embedding and sin/cos modulo embeddings for 100 residues (moduli
$2$--$101$), fused via FiLM. Three prediction heads (magnitude regression,
sign classification, and modulo prediction for 100 moduli) are trained jointly
on 274,705 OEIS sequences.
At the Large scale (91.5M parameters), IntSeqBERT achieves 95.85\% magnitude accuracy and
50.38\% Mean Modulo Accuracy (MMA) on the test set, outperforming a standard tokenised
Transformer baseline by $+8.9$\,pt and $+4.5$\,pt, respectively. An ablation removing the
modulo stream confirms it accounts for $+15.2$\,pt of the MMA gain and contributes an
additional $+6.2$\,pt to magnitude accuracy. A probabilistic Chinese Remainder Theorem (CRT)-based Solver converts the model's
predictions into concrete integers, yielding a 7.4-fold improvement in next-term prediction
over the tokenised-Transformer baseline (Top-1: 19.09\% vs.\ 2.59\%). Modulo spectrum analysis reveals a strong negative correlation
between Normalised Information Gain (NIG) and Euler's totient ratio $\varphi(m)/m$
($r = -0.851$, $p < 10^{-28}$), providing empirical evidence that composite moduli capture
OEIS arithmetic structure more efficiently via CRT aggregation.

\end{abstract}



\section{Introduction}
\label{sec:introduction}

The On-Line Encyclopedia of Integer Sequences (OEIS)~\cite{sloane1996}, as of January 2026,
catalogues 391,710 entries spanning combinatorics, number theory, algebra, and many
other branches of mathematics, making it the \emph{de facto} standard reference for integer
sequences. Each entry associates a finite integer sequence with its mathematical definition,
rendering the OEIS a uniquely machine-readable corpus of mathematical knowledge.

The long-term goal of this research is an AI system that discovers hidden number-theoretic
similarities between integer sequences arising in different mathematical fields or from
different algorithms, and thereby supports the generation of mathematical conjectures---a
goal in the same spirit as recent AI-driven discovery systems such as
AlphaProof~\cite{alphaproof2024}, the Ramanujan Machine~\cite{raayoni2021ramanujan}, and
FunSearch~\cite{romera2023funsearch} (Section~\ref{sec:background:ai-math}). A prerequisite
for this goal is a machine representation that captures the number-theoretic structure of
OEIS sequences. This paper takes a first step: we propose such a representation and measure,
through a masked-prediction task, how much arithmetic structure it actually captures.

The task we formalise is \emph{masked sequence modelling}: a random subset of positions in a
sequence is masked and the model is trained to predict each masked value from its surrounding
context. This task directly probes how well a model has internalised the arithmetic and
combinatorial laws governing integer sequences; next-term prediction is treated as a special
case and evaluated via a dedicated Solver component. The prior benchmark FACT~\cite{zurich-fact}
defines and evaluates five tasks including unmasking, establishing what a plain Transformer can
achieve, but it faces fundamental limitations in handling large integer values outside its fixed
token vocabulary and in learning multiplicative structure.

The principal difficulty of this task is the extreme heterogeneity of integer sequences.
Values range from single-digit constants to astronomically large factorials and exponential
sequences, with differences of tens of orders of magnitude within a single training corpus.
At the same time, many sequences obey \emph{residue constraints}---parity, combinatorial
patterns, or periodic remainders---that are independent of the magnitude of individual values.
The standard tokenisation approach of assigning each integer a discrete vocabulary token is
fundamentally ill-suited to this setting: it cannot represent integers outside its fixed token
vocabulary, embeds arithmetic structure in opaque token IDs, and breaks down in scale for large numbers.

We propose \textbf{IntSeqBERT}, a Transformer encoder pre-trained on the OEIS corpus via masked
sequence modelling, which overcomes the above challenges through a \emph{dual-stream} input
representation. Rather than tokenising integers, IntSeqBERT encodes each element along two
complementary axes:
\begin{itemize}
  \item \textbf{Magnitude stream}: a continuous log\-scale embedding of the
        absolute value, capturing growth behaviour and scale.
  \item \textbf{Modulo stream}: sin/cos embeddings for 100 residues modulo $2$ through $101$,
        capturing periodicity and number-theoretic structure.
\end{itemize}
The two streams are fused via FiLM (Feature-wise Linear Modulation)~\cite{perez2018film}.
During training, three predictors are jointly optimised in a multi-task objective: magnitude
regression, sign classification, and modulo prediction for 100 moduli. Concrete integer values
are recovered from the masked-position predictions (magnitude, sign, residue distributions)
using a probabilistic Chinese Remainder Theorem (CRT)-based \textbf{Solver} (Section~\ref{sec:solver}).

This combination of the modulo-spectrum representation and the CRT solver is a
\emph{neural-symbolic} design: residue arithmetic and the Chinese Remainder Theorem are
built into the model and the decoding procedure as explicit number-theoretic structure,
rather than left for the network to discover end-to-end. Our contribution is therefore not
the prediction accuracy itself---predicting integer sequences via machine learning is a
long-standing task, e.g.\ the Kaggle Integer Sequence Learning
competition~\cite{kaggle-integer-seq}---but the representation that explicitly encodes
number-theoretic structure, together with quantitative evidence that this representation
captures the structure of OEIS sequences. At the same time, our results document how
difficult direct neural prediction of OEIS terms remains even with this structure built in,
providing motivation for hybrid symbolic--neural approaches (Section~\ref{sec:conclusion}).

We evaluate IntSeqBERT against two baselines on 274,705 OEIS sequences across three model
sizes (Small: 6.4M, Middle: 29.0M, Large: 91.5M parameters), all on a
single GeForce RTX 3070 Ti (8\,GB VRAM).
The implementation, preprocessing scripts, and evaluation code are publicly
available.\footnote{\url{https://github.com/aabaa/IntSeqBERT}}

\noindent\textbf{Contributions.}
\begin{enumerate}
  \item \textit{IntSeqBERT architecture}: a dual-stream Transformer fusing magnitude and
        modular-arithmetic embeddings via FiLM, jointly trained on OEIS with magnitude
        regression, sign classification, and 100-dimensional modulo prediction. At the
        Large scale, IntSeqBERT achieves \textbf{95.85\%} magnitude accuracy and
        \textbf{50.38\%} MMA, surpassing Vanilla by $+8.9$\,pt and $+4.5$\,pt, respectively.
        The dual-stream representation yields a \textbf{7.4-fold} improvement in
        Solver-based next-term prediction (Top-1: 19.09\% vs.\ 2.59\%).

  \item \textit{Number-theoretic finding}: modulo spectrum analysis reveals that NIG is
        strongly negatively correlated with $\varphi(m)/m$ (where $\varphi$ denotes Euler's totient function; Pearson correlation of $r = {-}0.851$ ($p < 10^{-28}$)),
        providing quantitative evidence that composite moduli aggregate arithmetic structure
        more efficiently via CRT.

  \item \textit{Scaling behaviour}: modulo accuracy and Solver accuracy improve more steeply
        with model size than magnitude accuracy, suggesting arithmetic reasoning benefits
        disproportionately from increased capacity.
\end{enumerate}

\noindent\textbf{Paper organisation.}
Sections~\ref{sec:background}--\ref{sec:setup} review related work, present the architecture,
and describe the experimental setup. Section~\ref{sec:experiments} reports results, Section~\ref{sec:analysis} provides analyses, and Section~\ref{sec:conclusion} concludes.


\section{Background and Related Work}
\label{sec:background}

\subsection{AI for Mathematics}
\label{sec:background:ai-math}

Throughout mathematics, numerical pattern observation has seeded important
conjectures---from Monstrous Moonshine~\cite{conway1979monstrous} and the
BSD conjecture~\cite{birch1965} to Wiles's proof of the Shimura--Taniyama conjecture
for semistable elliptic curves, which resolved Fermat's Last Theorem~\cite{wiles1995},
and Candelas et al.'s mirror-symmetry prediction of 317,206,375
rational curves~\cite{candelas1991}.
Recent AI systems including AlphaGeometry~\cite{trinh2024alphageometry},
AlphaProof~\cite{alphaproof2024}, the Ramanujan Machine~\cite{raayoni2021ramanujan}, and
FunSearch~\cite{romera2023funsearch} demonstrate the viability of AI-driven mathematical
discovery. On the formalisation front, Urban's large-scale autoformalization
work~\cite{urban-autoformalization} shows that natural-language mathematics can be
automatically translated into machine-verifiable form.
Yet extracting arithmetic laws from integer sequences at scale remains largely
unexplored. This work aims to establish representational foundations for that capability.

\subsection{AI Research on OEIS}
\label{sec:background:oeis}

Program-synthesis approaches to OEIS include Alien Coding~\cite{urban-alien-coding},
which searches for the shortest reproducing code; QSynt~\cite{gauthier2023qsynt}, which
synthesises programs for 43,516 sequences via self-learning tree search; and Learning
Conjecturing~\cite{gauthier2025conjecturing}, which automatically proves 5,565 arithmetic problems.
FACT~\cite{zurich-fact} (Bel\v{c}\'{a}k et al., NeurIPS 2022) constructs a comprehensive OEIS
benchmark covering five tasks in order of difficulty---classification, similarity,
next sequence-part prediction, continuation, and
\emph{unmasking} (identified as the hardest).
Our Vanilla baseline corresponds to FACT's plain Transformer; IntSeqBERT extends it with
modulo feature engineering and FiLM fusion to overcome FACT's limitations with large integers
and multiplicative structure. The Solver-based next-term prediction evaluated in
Section~\ref{sec:solver-eval} corresponds to FACT's continuation task.

\subsection{Modulo Spectra as Arithmetic Feature Engineering}
\label{sec:background:modulo-engineering}

Deep networks exhibit a well-known \emph{spectral bias}~\cite{rahaman2019spectral}:
they learn low-frequency functions before high-frequency ones. In the context of integer
sequences, this is particularly problematic for multiplicatively growing sequences
(e.g., $n^2$, $n!$), where learning the growth law from token sequences alone requires
many layers.
This work mitigates the problem through feature engineering: because the results of
multiplication appear naturally in modular residues, supplying the modulo spectrum as explicit
input features reduces the network depth required to ``rediscover'' multiplicative structure.

More precisely, most OEIS sequences are generated by finite algorithms combining addition,
multiplication, and modular operations. Since modular arithmetic is a ring homomorphism over
$\mathbb{Z}$, the residue sequence $(x_i \bmod m)$ faithfully preserves additive and
multiplicative structure as a compact representation, regardless of the magnitudes involved.
Fermat's little theorem implies that $(a^n \bmod p)$ is periodic with period dividing $p-1$
for prime $p$ with $\gcd(a,p)=1$. Gauss's law of quadratic
reciprocity~\cite{gauss-reciprocity} and its generalisations further link residue information
across different primes.
Composite moduli simultaneously retain information from all prime power factors via CRT,
motivating the use of both prime and composite moduli in our spectrum.
Our modulo features cover $m \in \{2, 3, \ldots, 101\}$ (100 moduli), capturing power-residue
structure broadly across primes and composites.

\subsection{Positioning and Neural Representations}
\label{sec:background:repr-position}

Tokenisation~(the LLM mainstream) cannot handle out-of-vocabulary values and buries
arithmetic structure in opaque token IDs. Log-scale continuous embeddings improve scale
invariance but miss arithmetic periodicity. We instead repurpose sinusoidal
embeddings~\cite{vaswani2017attention} from positions to \emph{modular residues of values},
and apply FiLM~\cite{perez2018film} to modulate the magnitude stream via the modulo stream.

In contrast to the Gauthier and Urban group's program-synthesis line~\cite{urban-alien-coding,gauthier2023qsynt,gauthier2025conjecturing},
which explicitly describes each sequence's generation rule, this work uses \emph{representation
learning} to acquire shared arithmetic structure across the OEIS corpus.
We view the two approaches as complementary: program synthesis seeks a formula or program
that generates a given sequence, whereas our representations aim to judge whether two
sequences are number-theoretically similar even when they are generated in entirely
different ways. Pre-trained IntSeqBERT embeddings could in turn serve as a search heuristic
for program synthesis, or as an index for retrieving similar sequences.
Where FACT~\cite{zurich-fact} identified the limits of a plain tokenised Transformer,
IntSeqBERT overcomes them via modulo-spectrum feature engineering and FiLM fusion.


\section{IntSeqBERT}
\label{sec:model}

\subsection{Problem Formulation}
\label{sec:formulation}

Let $\mathbf{x} = (x_1, x_2, \ldots, x_L)$ with $x_i \in \mathbb{Z}$ and $L \leq 128$ be a
finite prefix of an OEIS sequence. We adopt \emph{masked sequence modelling}: a random subset
of positions is masked and the model is trained to predict each masked value. Specifically, for
each masked position $i$, three quantities are predicted:
\begin{enumerate}
  \item \textbf{Magnitude}: $v_i =
    \begin{cases} 0 & (x_i = 0),\\ 1 + \log_{10}|x_i| & (x_i \neq 0) \end{cases} \in \mathbb{R}_{\geq 0}$
  \item \textbf{Sign}: $s_i \in \{+, -, 0\}$ (3-class label)
  \item \textbf{Residues}: $r_i^{(m)} = x_i \bmod m$ for each $m \in \{2, 3, \ldots, 101\}$
        (100 independent classification targets)
\end{enumerate}
This decomposition separates magnitude, sign, and periodic arithmetic structure into
complementary supervision signals.

\subsection{Dual-Stream Representation}
\label{sec:dualstream}

Two feature vectors are computed for each element $x_i$ prior to learnable embeddings.

\textbf{Magnitude features} $\mathbf{f}_i^{\text{mag}} \in \mathbb{R}^4$:
\[
  \mathbf{f}_i^{\text{mag}} = \bigl[v_i,\;\mathbf{1}[x_i > 0],\;\mathbf{1}[x_i < 0],\;\mathbf{1}[x_i = 0]\bigr].
\]
The last three components are a one-hot sign representation. For astronomically large integers
exceeding the \texttt{float64} range, $|x_i|$ is replaced by its decimal digit count.

\textbf{Modulo features} $\mathbf{f}_i^{\text{mod}} \in \mathbb{R}^{200}$:
For each modulus $m \in \{2, \ldots, 101\}$, let $r = x_i \bmod m$, and embed the residue as
a point on the unit circle:
\[
  \phi_m(r) = \left[\sin\!\left(\tfrac{2\pi r}{m}\right),\;\cos\!\left(\tfrac{2\pi r}{m}\right)\right] \in \mathbb{R}^2.
\]
Concatenating all 100 moduli gives $\mathbf{f}_i^{\text{mod}} \in \mathbb{R}^{200}$.
This sin/cos embedding is equivariant to the group structure of $\mathbb{Z}/m\mathbb{Z}$ and
avoids discontinuities at wrap-around boundaries.

The two feature vectors are projected to the model hidden dimension $d$ by independent
projection layers (two-layer MLP for magnitude; affine for modulo):
\[
  \mathbf{h}_i^{\text{mag}} = \mathrm{MLP}_{\text{mag}}\!\left(\mathbf{f}_i^{\text{mag}}\right),
  \quad
  \mathbf{h}_i^{\text{mod}} = W_{\text{mod}}\,\mathbf{f}_i^{\text{mod}} + \mathbf{b}_{\text{mod}},
  \quad \mathbf{h}_i^{\text{mag}},\,\mathbf{h}_i^{\text{mod}} \in \mathbb{R}^d.
\]

The two streams are fused by FiLM~\cite{perez2018film}: the modulo embedding generates
element-wise scale $\boldsymbol{\gamma}_i$ and shift $\boldsymbol{\beta}_i$:
\[
  \boldsymbol{\gamma}_i = W_\gamma\,\mathbf{h}_i^{\text{mod}},
  \quad
  \boldsymbol{\beta}_i = W_\beta\,\mathbf{h}_i^{\text{mod}},
  \quad
  \mathbf{e}_i = (1 + \boldsymbol{\gamma}_i) \odot \mathbf{h}_i^{\text{mag}} + \boldsymbol{\beta}_i.
\]
A ReLU is applied after the modulo projection, and dropout is inserted before FiLM.
Standard sinusoidal positional encodings are added to $\mathbf{e}_i$ before the encoder.

\begin{figure}[htbp]
  \centering
  \includegraphics[width=\linewidth]{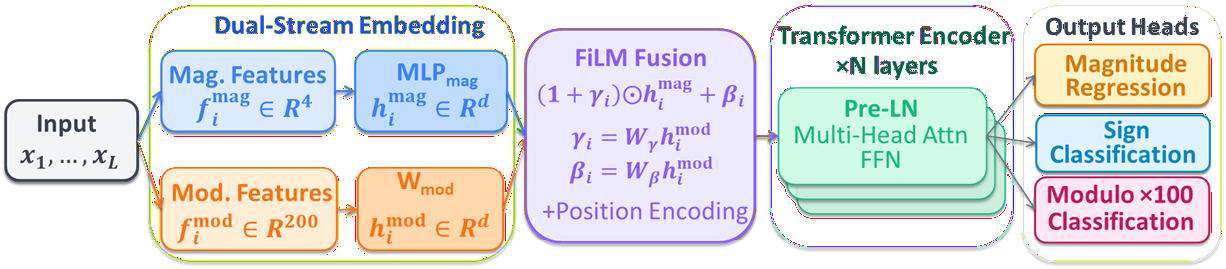}
  \caption{IntSeqBERT architecture. The Dual-Stream Embedding block projects
           $\mathbf{f}_i^{\mathrm{mag}}\!\in\!\mathbb{R}^4$ and
           $\mathbf{f}_i^{\mathrm{mod}}\!\in\!\mathbb{R}^{200}$ to $\mathbb{R}^d$ via
           $\mathrm{MLP}_{\mathrm{mag}}$ and $W_{\mathrm{mod}}$, then fuses them with FiLM:
           $\mathbf{e}_i = (1+\boldsymbol{\gamma}_i)\odot\mathbf{h}_i^{\mathrm{mag}}+\boldsymbol{\beta}_i$.
           Positional encodings are added before the Pre-LN Transformer encoder.
           Three prediction heads produce
           $\hat{v},\log\hat{\sigma}^2\!\in\!\mathbb{R}$ (magnitude regression),
           $\hat{s}\!\in\!\{+,-,0\}$ (sign classification), and
           $\hat{r}^{(m)}\!\in\!\{0,\ldots,m{-}1\}$ ($100\times$ modulo classification).}
  \label{fig:architecture}
\end{figure}

The overall architecture is illustrated in Fig.~\ref{fig:architecture}.

\subsection{Encoder, Heads, and Training}
\label{sec:encoder}

The fused sequence $(\mathbf{e}_1, \ldots, \mathbf{e}_L)$ is processed by a standard
Transformer encoder~\cite{vaswani2017attention} with Pre-Layer Normalisation~\cite{xiong2020layer}.
Three model sizes are evaluated (Table~\ref{tab:models}).

\begin{table}[htbp]
\caption{Model configurations.}
\label{tab:models}
\centering
\begin{tabular}{lcccr}
\hline
Config  & Layers & $d$ & Heads & Parameters \\
\hline
Small   & 6      & 256 & 4     & 6.4M       \\
Middle  & 8      & 512 & 8     & 29.0M      \\
Large   & 12     & 768 & 12    & 91.5M      \\
\hline
\end{tabular}
\end{table}

Let $\mathbf{z}_i \in \mathbb{R}^d$ be the encoder output at masked position $i$.
Three prediction heads are applied: (i)~\textbf{Magnitude head} --- a two-layer MLP
($d \to d \to 2$, ReLU) outputting $(\mu_i, \log\sigma_i^2)$, predicting $\hat{v}_i = \mu_i$
(a log-scale estimate of $v_i$; $\sigma_i^2 = \exp(\log\sigma_i^2)$ is used by the Solver);
(ii)~\textbf{Sign head} --- $\hat{s}_i = \operatorname{softmax}(W_{\text{sign}}\mathbf{z}_i)$;
(iii)~\textbf{Modulo head} --- 100 independent linear classifiers, one per modulus, with total
output dimension $\sum_{m=2}^{101}m = 5{,}150$.

The multi-task loss is $\mathcal{L} = \mathcal{L}_{\text{mag}} + \mathcal{L}_{\text{sign}} + 2\,\mathcal{L}_{\text{mod}}$.
The weight~2 on $\mathcal{L}_{\text{mod}}$ is intentionally set higher to emphasise the
newly introduced modulo stream; adaptive weighting was attempted but caused training instability.
$\mathcal{L}_{\text{mag}}$ is Huber loss; $\mathcal{L}_{\text{sign}}$ is cross-entropy over
three sign classes; $\mathcal{L}_{\text{mod}} = \frac{1}{100}\sum_{m=2}^{101}
\frac{1}{\ln m}\mathcal{L}_{\text{CE}}^{(m)}$, where $\mathcal{L}_{\text{CE}}^{(m)}$ is the
cross-entropy loss of the $m$-class classifier for modulus $m$, normalised by $\ln m$
(the maximum entropy of a uniform distribution over $m$ classes) to compensate for varying class counts.
All losses are computed only at masked positions.

\subsection{Baselines}
\label{sec:baselines}

\textbf{Vanilla Transformer} maps each integer to a token ID in a vocabulary of 20,003 entries
(non-negative values 0--19{,}999 plus \texttt{PAD}, \texttt{MASK}, \texttt{UNK}); negative
integers and values $\geq 20{,}000$ are replaced by \texttt{UNK}. The same three prediction heads are applied to the token
embeddings. The vocabulary size 20,003 was chosen so that memory consumption matches
IntSeqBERT under the 8\,GB VRAM constraint. The prior benchmark FACT~\cite{zurich-fact}
handles values up to several million, presupposing larger compute resources than our setting.

\textbf{Ablation (magnitude-only)} is identical to IntSeqBERT but uses only the magnitude
stream; the FiLM module is removed and $\mathbf{e}_i = \mathbf{h}_i^{\text{mag}}$. This
isolates the contribution of the modulo stream.

\subsection{Solver}
\label{sec:solver}

The pre-trained model outputs magnitude $(\mu, \log \sigma^2)$, sign, and modulo
distributions at masked positions. \textbf{Solver} recovers concrete integers from these
predictions.

The Solver derives the $3\sigma$ interval $[n_{\min}, n_{\max}]$ from the magnitude prediction
and dynamically selects one of three modes based on the search width
$\Delta n = |n_{\max} - n_{\min}|$ (Table~\ref{tab:solver}).

\begin{table}[htbp]
\caption{Solver operating modes.}
\label{tab:solver}
\centering
\begin{tabular}{llp{5.5cm}}
\hline
Mode     & Condition                              & Method \\
\hline
Dense    & $\Delta n \leq 10^6$                   & Enumerate all integers in range \\
Sieve    & $10^6 < \Delta n \leq 10^{14}$         & CRT beam search anchored on high-confidence moduli \\
CRT      & $\Delta n > 10^{14}$                   & Sparse CRT beam search to generate large integers directly \\
\hline
\end{tabular}
\end{table}

When the sign prediction is zero, the Solver immediately returns 0. If no valid candidate is
found in the search range, the call is recorded as \texttt{none}.

Each candidate $n$ is scored by $\text{score}(n) = \alpha_{\text{mag}} + 0.3\,\alpha_{\text{mod}}$,
where $\alpha_{\text{mag}} = -(v_n-\mu_i)^2/(2\sigma_i^2)$ and
$\alpha_{\text{mod}} = \sum_{m}\ln P(n\bmod m)$.
The coefficient 0.3 approximates the proportion of primes between 2 and 101
($26/100 = 0.26$), compensating for the information overlap among composite moduli
that share prime factors.
The top $k$ candidates are returned and evaluated as Solver Top-$k$ (Section~\ref{sec:solver-eval}).


\section{Dataset and Experimental Setup}
\label{sec:setup}

\subsection{Dataset}
\label{sec:dataset}

As of January 2026, OEIS contains 391,710 sequences. We exclude sequences with fewer than ten
terms and those carrying any of eleven incompatible OEIS tags (out-of-scope representations:
\texttt{cons}/\texttt{cofr}/\texttt{frac}/\texttt{base}/\texttt{word}; undefined task:
\texttt{fini}/\texttt{tabl}; quality: \texttt{dead}/\texttt{unkn}/\texttt{less}/\texttt{dumb}),
yielding 274,705 sequences (70.1\%) as our standard dataset (\texttt{std}).
The intent of this filtering is to narrow the corpus to sequences whose number-theoretic
structure can be studied through the integer values themselves: the excluded representations
depend strongly on a particular encoding (e.g.\ decimal digit expansions), and the
quality-label tags mark incomplete or deprecated entries. We acknowledge that some of these
choices are not strictly necessary: the quality labels \texttt{less} and \texttt{dumb} are
based on subjective criteria, and \texttt{tabl} sequences can be encoded as one-dimensional
sequences by a standard transformation. These exclusions are a practical decision for
constructing a first benchmark; designing a more systematic filtering criterion---for
example, classifying sequences by the complexity of their generation rule or by their
prediction difficulty---is left for future work.

The corpus is split 8:1:1 (seed~42): 219,765 training / 27,470 validation / 27,470 test sequences.

Each sample corresponds to a sequence prefix of at most $L = 128$ terms (longer sequences are
truncated). In the raw test corpus (before truncation), sequence lengths range from 10 to 168
terms (mean 42.5). After truncation, all model inputs satisfy $L \le 128$, while still ensuring
that the Solver receives at least 9 terms of preceding context (Section~\ref{sec:solver-eval}).

For scale-stratified evaluation, we define five magnitude buckets by value range:

\begin{table}[htbp]
\caption{Magnitude buckets used for scale-stratified evaluation.}
\label{tab:buckets}
\centering
\begin{tabular}{lll}
\hline
Bucket       & Value range                     & Element fraction \\
\hline
Small        & $|x| < 10^2$                    & 52.7\%           \\
Medium       & $10^2 \leq |x| < 10^5$          & 29.9\%           \\
Large        & $10^5 \leq |x| < 10^{20}$       & 16.2\%           \\
Huge         & $10^{20} \leq |x| < 10^{50}$    &  1.1\%           \\
Astronomical & $|x| \geq 10^{50}$              &  0.0\%           \\
\hline
\end{tabular}
\end{table}

\textbf{Limitations of the split.} The random split may place related sequences (e.g.,
Fibonacci and Lucas) in different splits; family-aware splits are left for future work.

\subsection{Training Configuration}
\label{sec:training-config}

All models share the hyperparameters shown in Table~\ref{tab:hyperparams}.
Some OEIS values reach $|x| \approx 10^{210}$, causing FP16 overflow; all computation is
therefore conducted in FP32. All results are reported by applying the model trained for the
fixed 200 epochs to the test set.

\begin{table}[htbp]
\caption{Training hyperparameters (common to all models).}
\label{tab:hyperparams}
\centering
\begin{tabular}{ll}
\hline
Hyperparameter        & Value \\
\hline
Epochs                & 200 (no early stopping) \\
Batch size            & 32 (gradient accumulation 2 steps; effective 64) \\
Learning rate         & $5 \times 10^{-5}$ \\
Warmup fraction       & 10\% \\
Optimiser             & AdamW, weight decay $0.01$ \\
Numerical precision   & FP32 (AMP disabled) \\
Mask probability      & 0.15 \\
GPU                   & GeForce RTX 3070 Ti (8\,GB VRAM) $\times$ 1 \\
\hline
\end{tabular}
\end{table}

\subsection{Evaluation Metrics}
\label{sec:metrics}

We evaluate using the following metrics:
\textbf{Magnitude Accuracy} (\textbf{Mag Acc}, $\text{Acc}_{0.5}$): fraction of masked positions with
$|\hat{v}_i - v_i| < 0.5$ ($\sqrt{10}{\approx}3.16\times$ tolerance in integer scale).
This metric does \emph{not} measure exact integer match; it measures whether the magnitude
prediction is precise enough to set a realistic search range for the Solver's Dense and
Sieve modes (Section~\ref{sec:solver}). Concretely, if the true value is $10^6$, any
prediction between roughly $3.16 \times 10^5$ and $3.16 \times 10^6$ counts as correct.
Exact integer match is evaluated separately by Solver Top-$k$ accuracy.
\textbf{Sign Accuracy} (\textbf{Sign Acc}): fraction with correct predicted sign class.
\textbf{Mean Modulo Accuracy} (\textbf{MMA}): mean classification accuracy across all 100 moduli.
\textbf{Solver Top-$k$}: fraction of next-term instances where the true value appears in the
Solver's top-$k$ candidates (Section~\ref{sec:solver-eval}).
\textbf{Normalised Information Gain} (\textbf{NIG}): $1 - \mathcal{L}_{\text{CE}}^{(m)} / \ln m$
for modulus $m$; higher NIG indicates stronger learned periodic structure.


\section{Experiments}
\label{sec:experiments}

\subsection{Main Results}
\label{sec:main-results}

Table~\ref{tab:main} shows test-set performance for all model sizes and variants. IntSeqBERT
consistently outperforms both baselines on all scales and metrics. At the Large scale,
IntSeqBERT surpasses the Vanilla baseline by $+8.9$\,pt in Mag Acc and $+4.5$\,pt in MMA.
The Ablation model shows the largest MMA drop at Large scale ($-15.2$\,pt), directly
quantifying the contribution of the arithmetic-periodicity features. Notably, the Ablation
maintains non-trivial Mag Acc, but the remaining gap confirms that modulo information contributes
meaningfully to magnitude regression as well, while being essential for modulo prediction.

\begin{table}[htbp]
\caption{Test results. Mag Acc (\%), Sign Acc (\%), MMA (\%). \textbf{Bold} indicates the best value within each size group.}
\label{tab:main}
\centering
\begin{tabular}{llrrr}
\hline
Size   & Model          & Mag Acc        & Sign Acc       & MMA            \\
\hline
Small  & \textbf{IntSeq}& \textbf{94.73} & \textbf{97.78} & \textbf{40.43} \\
Small  & Vanilla        & 85.73          & 96.91          & 36.21          \\
Small  & Ablation       & 93.72          & 97.39          & 25.97          \\
Middle & \textbf{IntSeq}& \textbf{95.71} & \textbf{98.34} & \textbf{46.88} \\
Middle & Vanilla        & 87.37          & 97.42          & 42.53          \\
Middle & Ablation       & 92.45          & 97.90          & 31.93          \\
Large  & \textbf{IntSeq}& \textbf{95.85} & \textbf{98.54} & \textbf{50.38} \\
Large  & Vanilla        & 86.97          & 97.66          & 45.85          \\
Large  & Ablation       & 89.70          & 98.29          & 35.22          \\
\hline
\end{tabular}
\end{table}

\begin{figure}[htbp]
  \centering
  \includegraphics[width=\linewidth]{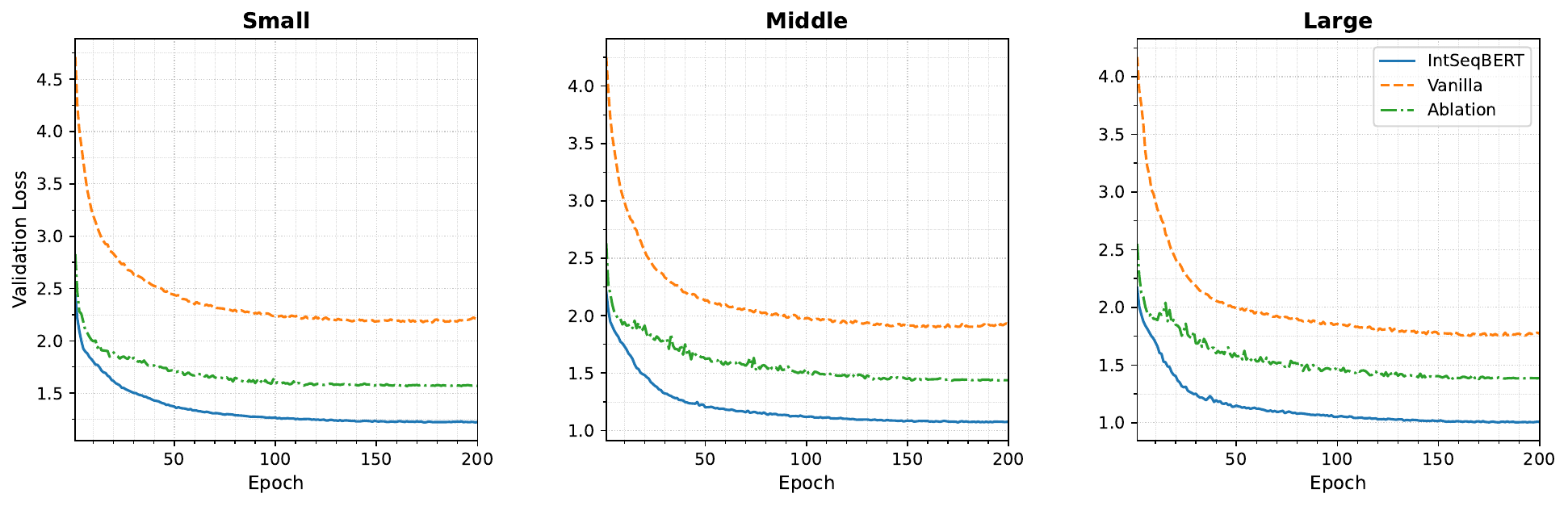}
  \caption{Validation loss learning curves for all scales (Small / Middle / Large) and all
           variants. IntSeqBERT (solid blue) consistently achieves lower loss than Vanilla
           (dashed orange) and Ablation (dash-dot green). At the Large scale, IntSeqBERT
           converges to Val Loss = 1.01 at epoch 200.}
  \label{fig:learning_curve}
\end{figure}

\textbf{Learning curves.} Fig.~\ref{fig:learning_curve} shows validation loss across all
variants. Large IntSeqBERT converges from 2.17 to 1.01 over 200 epochs with no overfitting
(Train~1.00 / Val~1.01). Vanilla (1.77) and Ablation (1.39) maintain higher final loss.

\subsection{Magnitude Prediction}
\label{sec:magnitude}

\textbf{Scale-stratified analysis.} Table~\ref{tab:scale} reports per-bucket MSE on the test
set for Large models. The Vanilla model suffers catastrophic degradation in the Large bucket
(MSE = 2.10, $13\times$ that of IntSeqBERT), caused by all out-of-vocabulary integers being
absorbed into the \texttt{UNK} token. IntSeqBERT achieves the best MSE in all buckets except
Small, with particularly pronounced advantage at Medium and above. The Ablation degrades in the
Medium bucket without modulo context (MSE = 0.116 vs.\ IntSeqBERT 0.051), and collapses more
severely in the Huge and Astronomical buckets, indicating that FiLM modulation by the modulo
stream acts as an arithmetic structural constraint on magnitude estimation for large integers.

\begin{table}[htbp]
\caption{Scale-stratified MSE on the test set (Large models). Lower is better.}
\label{tab:scale}
\centering
\begin{tabular}{lrrr}
\hline
Bucket        & IntSeq         & Vanilla & Ablation \\
\hline
Small         & 0.111          & 0.138   & \textbf{0.103} \\
Medium        & \textbf{0.051} & 0.071   & 0.116   \\
Large         & \textbf{0.162} & 2.100   & 0.381   \\
Huge          & \textbf{2.082} & 22.73   & 5.021   \\
Astronomical  & \textbf{110.4} & 840.0   & 532.6   \\
\hline
\end{tabular}
\end{table}

\begin{figure}[htbp]
  \centering
  \includegraphics[width=\linewidth]{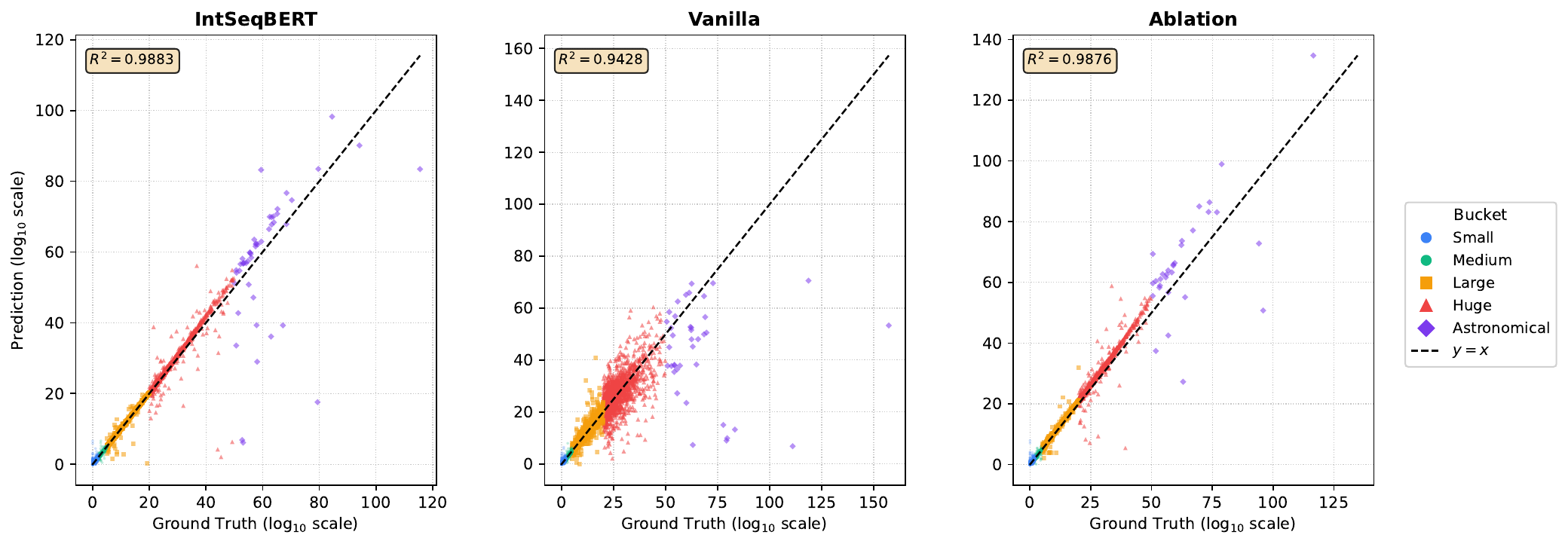}
  \caption{Predicted vs.\ true magnitude ($\log_{10}$ scale) for Large models. Points are
           coloured by bucket (Small=blue circle, Medium=green circle, Large=yellow-orange
           square, Huge=red triangle, Astronomical=purple diamond). IntSeqBERT achieves
           $R^2 = 0.988$ vs.\ Vanilla $R^2 = 0.943$; Vanilla shows pronounced scatter above
           the Large bucket.}
  \label{fig:scatter}
\end{figure}

Fig.~\ref{fig:scatter} shows predicted vs.\ true magnitude scatter plots for the three Large
model variants. IntSeqBERT achieves the highest coefficient of determination ($R^2 = 0.988$),
substantially reducing off-diagonal dispersion in the Large and Huge buckets compared to
Vanilla ($R^2 = 0.943$), visually corroborating the MSE results of Table~\ref{tab:scale}.

\subsection{Modulo Spectrum Analysis}
\label{sec:modulo-spectrum}

\begin{figure}[h]
  \centering
  \includegraphics[width=\linewidth]{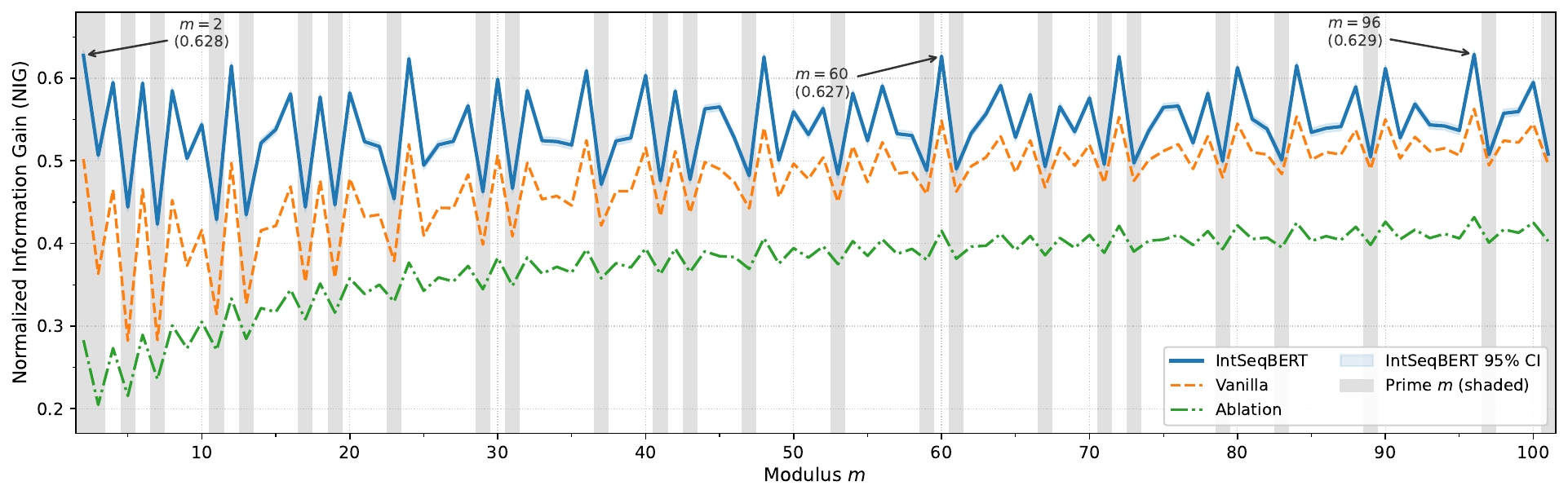}
  \caption{NIG spectrum for moduli $m = 2, \ldots, 101$ (Large models). Grey shading marks
           prime moduli. The 95\% CI for IntSeqBERT (light blue band) is computed by
           bootstrapping.}
  \label{fig:nig_spectrum}
\end{figure}

Fig.~\ref{fig:nig_spectrum} shows the NIG spectrum for $m = 2, \ldots, 101$ using Large
models. IntSeqBERT (solid blue) exceeds both Vanilla and Ablation across the entire spectrum,
with a visually clear contrast between prime and composite moduli.

\begin{figure}[htbp]
  \centering
  \includegraphics[width=0.72\linewidth]{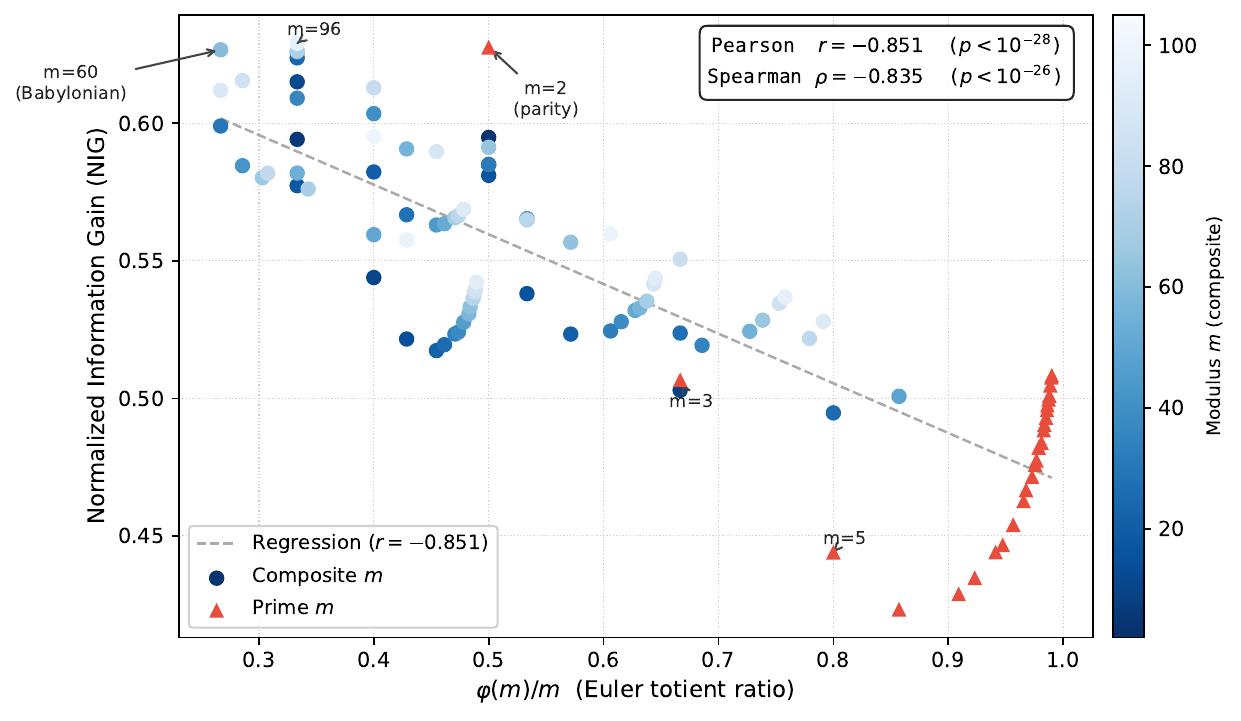}
  \caption{NIG vs.\ Euler's totient ratio $\varphi(m)/m$ (Large IntSeqBERT). Composite
           moduli (blue circles, shade proportional to $m$) and prime moduli (red triangles)
           are shown separately. The regression line (grey dashed) indicates Pearson correlation of
           $r = -0.851$ ($p < 10^{-28}$). Notable moduli $m = 2$ (parity),
           $m = 60$ (Babylonian number), and $m = 96$ ($2^5 \times 3$ composite) are annotated.}
  \label{fig:nig_phi}
\end{figure}

\textbf{Finding 1: NIG is strongly negatively correlated with Euler's totient ratio.}
A Pearson correlation of $r = -0.851$ ($p < 10^{-28}$) is observed between NIG and
$\varphi(m)/m = \prod_{p \mid m}(1 - 1/p)$ (Fig.~\ref{fig:nig_phi}). Composite moduli with
many small prime factors achieve higher NIG, consistent with a CRT aggregation effect: when
$m$ is a common multiple of smaller moduli $m_1, m_2, \ldots$, then $x \bmod m$ encodes the
information of all those moduli simultaneously.
The highest NIG across all models and scales is achieved at $m = 96 = 2^5 \times 3$
($\varphi(96)/96 = 1/3$; Large IntSeq: NIG = 0.629, 95\% CI [0.622, 0.634]), consistent with
$m = 96$ being the largest modulus with totient ratio $1/3$ among $\{12, 24, 48, 72, 96\}$ and
thus distinguishing the widest value range. The exception is $m = 2$ (prime, NIG = 0.628,
second highest), reflecting the corpus-wide ubiquity of parity in OEIS sequences.

While this correlation is an empirical observation rather than a theorem, it carries a
design implication for the modulo stream itself: composite moduli reflect constraints from
several prime factors and prime powers at once, so the CRT aggregation effect lets a model
extract more number-theoretic information from fewer moduli. This gives a guideline for
choosing moduli---for example, when extending the spectrum beyond $m = 101$, adding
composite numbers with many small prime factors (such as highly composite numbers) is
likely to be more effective than simply adding more primes.

\textbf{Finding 2: Parity (mod 2) accuracy stratifies models.}
Mod-2 accuracy at the Large scale is 85.65\% (IntSeq), 81.40\% (Vanilla), and 72.13\%
(Ablation), making parity the single modulus most sensitive to the modulo stream ($-13.5$\,pt
when removed). Table~\ref{tab:moduli} lists representative modulus accuracies.

\begin{table}[h]
\caption{Representative modulus accuracies for Large models (\%).}
\label{tab:moduli}
\centering
\begin{tabular}{rrrrl}
\hline
$m$ & IntSeq & Vanilla & Ablation & Interpretation \\
\hline
  2 & 85.65  & 81.40   & 72.13    & Parity \\
  3 & 72.62  & 65.22   & 53.72    & Ternary residue \\
  5 & 60.37  & 50.07   & 42.63    & Last decimal digit modulo 5 \\
 10 & 58.38  & 49.25   & 39.47    & Least significant decimal digit \\
 60 & 53.97  & 47.87   & 35.12    & Babylonian number (highly composite) \\
 96 & 51.82  & 47.29   & 34.44    & Composite ($2^5 \times 3$) \\
100 & 48.51  & 45.60   & 33.51    & Centesimal residue \\
\hline
\end{tabular}
\end{table}

\subsection{Solver-Based Next-Term Prediction}
\label{sec:solver-eval}

The Solver reconstructs candidate integers from the model's magnitude, sign, and modulo
predictions and ranks them by likelihood. We evaluate exact-match accuracy on 10,000 test
samples, where each sample corresponds to one OEIS sequence and the target is the \emph{last
term} (all preceding terms are given as context).

\emph{Valid-candidate rate} is the fraction of samples for which the Solver returns at least
one candidate. Vanilla always returns a candidate from its vocabulary softmax (rate = 100\%).
IntSeqBERT and Ablation may fail to find a valid candidate within the search range
(\texttt{none} mode; 13.4\% of Large-IntSeq calls), reducing their valid-candidate rate.

Table~\ref{tab:solver-eval} shows Solver evaluation results. Large-scale IntSeqBERT achieves
a Top-1 accuracy of 19.09\%, a \textbf{7.4-fold} improvement over the Vanilla
baseline (2.59\%), counting \texttt{none} returns as misses.

\begin{table}[htbp]
\caption{Solver evaluation: Top-1 and Top-10 exact-match accuracy (\%) and valid-candidate rate.}
\label{tab:solver-eval}
\centering
\begin{tabular}{llrrrr}
\hline
Size   & Model           & Top-1          & Top-10         & Sign Acc       & Valid rate (\%) \\
\hline
Small  & \textbf{IntSeq} & \textbf{14.05} & \textbf{21.00} & \textbf{98.73} & 90.59 \\
Small  & Vanilla         & 2.43           & 3.24           & 92.92          & 100.0 \\
Small  & Ablation        & 7.42           & 17.33          & 98.50          & 90.17 \\
Middle & \textbf{IntSeq} & \textbf{17.02} & \textbf{22.62} & \textbf{99.02} & 86.31 \\
Middle & Vanilla         & 2.43           & 3.41           & 92.71          & 100.0 \\
Middle & Ablation        & 9.88           & 20.52          & 98.74          & 90.34 \\
Large  & \textbf{IntSeq} & \textbf{19.09} & \textbf{26.23} & \textbf{99.02} & 86.64 \\
Large  & Vanilla         & 2.59           & 3.80           & 92.05          & 100.0 \\
Large  & Ablation        & 11.75          & 21.79          & 98.94          & 86.99 \\
\hline
\end{tabular}
\end{table}

\textbf{Accuracy by magnitude bucket.} Table~\ref{tab:solver-bucket} compares Solver
Top-1 accuracy across buckets for all three Large models. IntSeqBERT achieves 68.34\%
Top-1 in the Small bucket, while the Ablation reaches 54.55\% and Vanilla only 14.11\%.
Vanilla collapses to 0\% for Medium and above due to the UNK-token degradation of its
magnitude predictions, whereas IntSeqBERT retains meaningful accuracy up to the Medium
bucket (20.82\%).

\begin{table}[htbp]
\caption{Solver Top-1 / Top-10 accuracy (\%) by magnitude bucket (Large models).
         \textbf{Bold}: best per bucket.}
\label{tab:solver-bucket}
\centering
\begin{tabular}{l rr rr rr}
\hline
 & \multicolumn{2}{c}{\textbf{IntSeqBERT}} & \multicolumn{2}{c}{Vanilla} & \multicolumn{2}{c}{Ablation} \\
Bucket & Top-1 & Top-10 & Top-1 & Top-10 & Top-1 & Top-10 \\
\hline
Small        & \textbf{68.34} & \textbf{88.50} & 14.11 & 20.71 & 54.55 & 86.92 \\
Medium       & \textbf{20.82} & \textbf{31.50} &  0.00 &  0.00 &  5.61 & 18.88 \\
Large        & \textbf{ 0.31} & \textbf{ 0.67} &  0.00 &  0.00 &  0.03 &  0.05 \\
Huge         & \textbf{ 0.09} & \textbf{ 0.18} &  0.00 &  0.00 &  0.00 &  0.00 \\
Astronomical &  0.00          &  0.00          &  0.00 &  0.00 &  0.00 &  0.00 \\
\hline
\end{tabular}
\end{table}

\textbf{Mode breakdown (Large IntSeqBERT).} Dense mode (24.0\% of calls) achieves the highest
Top-1 accuracy at 61.06\%; Sieve mode (36.7\%) achieves 5.36\%; the zero-prediction shortcut
(2.7\%) achieves 89.96\%. CRT mode (23.2\%) achieves near-zero accuracy (0.09\%); 13.4\% of
calls return no valid candidate. The CRT limitation is discussed in Section~\ref{sec:limitations}.


\section{Analysis and Discussion}
\label{sec:analysis}

\subsection{Ablation Study: Contribution of the Modulo Stream}
\label{sec:ablation}

To isolate the contribution of the modulo stream and FiLM fusion, we compare IntSeqBERT with
the magnitude-only Ablation model. As seen in Table~\ref{tab:main}, the modulo stream
delivers the largest gains in modulo prediction (MMA $+15.2$\,pt, parity $+13.5$\,pt at Large
scale). Its contribution to magnitude prediction is smaller but significant (Acc$_{0.5}$
$+6.2$\,pt, MSE $-0.228$), consistent with the intuition that knowing $x_i \bmod m$ for
$m \in \{2,\ldots,101\}$ substantially constrains the set of plausible magnitude values.
Solver accuracy improves by $+7.3$\,pt, primarily because more accurate residue information
refines scoring in Dense and Sieve modes.

\subsection{Scaling Behaviour}
\label{sec:scaling}

Table~\ref{tab:scale_trend} shows how IntSeqBERT's metrics improve with model size.
Magnitude accuracy improves only modestly from Small to Large ($+1.1$\,pt), whereas modulo
accuracy improves substantially ($+10.0$\,pt). This is consistent with the intuition that
modular arithmetic is more compositional and thus benefits more from increased representational
capacity. Solver Top-1 accuracy follows the modulo trend with consistent improvement ($+5.0$\,pt).

\begin{table}[htbp]
\caption{IntSeqBERT scaling: test-split metrics (Small / Middle / Large).}
\label{tab:scale_trend}
\centering
\begin{tabular}{lrrrr}
\hline
Metric              & Small  & Middle & Large  & $\Delta$ (S$\to$L) \\
\hline
Mag Acc (\%)        &  94.73 &  95.71 &  95.85 & $+1.12$ \\
MMA (\%)            &  40.43 &  46.88 &  50.38 & $+9.95$ \\
mod-2 accuracy (\%) &  81.97 &  84.50 &  85.65 & $+3.68$ \\
Solver Top-1 (\%)   &  14.05 &  17.02 &  19.09 & $+5.04$ \\
Magnitude MSE       &  0.228 &  0.164 &  0.142 & $-0.086$\\
\hline
\end{tabular}
\end{table}

\begin{figure}[htbp]
  \centering
  \includegraphics[width=\linewidth]{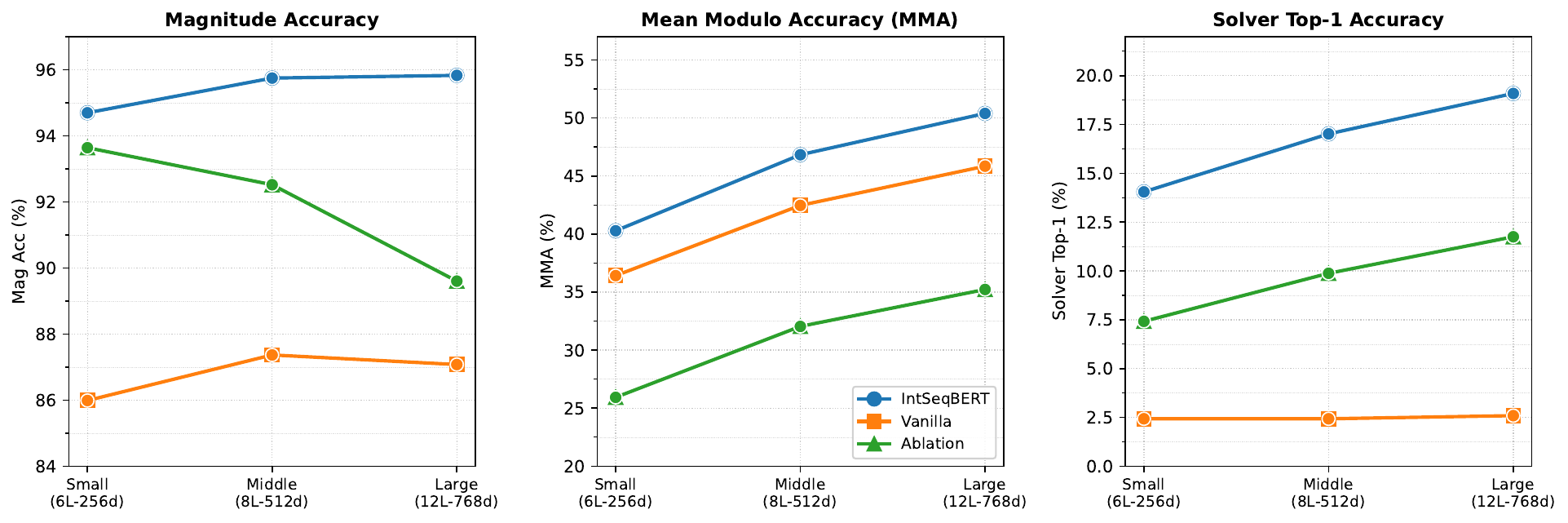}
  \caption{Scaling behaviour (Small / Middle / Large). Left: Mag Acc improves only $+1.1$\,pt.
           Centre: MMA for IntSeqBERT improves $+10.1$\,pt (40.3\%$\to$50.4\%).
           Right: Solver Top-1 improves $+5.0$\,pt, with a growing gap over Vanilla.}
  \label{fig:scaling}
\end{figure}

The trends in Table~\ref{tab:scale_trend} are visualised in Fig.~\ref{fig:scaling}, which
contrasts the differential scaling of modulo and magnitude accuracy across model sizes.

\subsection{Alternative Representations: Digit-Based Tokenisation}
\label{sec:digit-discussion}

A natural alternative to our handcrafted features is digit-based tokenisation, treating each
decimal digit as a separate token, possibly with digit-by-digit generation on the output
side. We have not made an experimental comparison with digit-level representations in this
paper. What our representation targets is not the pattern of decimal digits but the
number-theoretic structure of the integer itself: the modulo stream supplies the residues of
each value under 100 different moduli---information about modular arithmetic---to the model
directly, whereas the same information is only indirectly recoverable from a digit sequence.
Digit-by-digit generation is a reasonable alternative on the output side, but it changes the
problem setting from predicting an integer as a single value to the sequential prediction of
a digit string. A controlled comparison between the two settings is an interesting direction
for future work.

\subsection{Limitations}
\label{sec:limitations}

\textbf{Difficulty with large integers.} Solver accuracy collapses to near zero for
$|x| \geq 10^{20}$ (Huge and Astronomical buckets): CRT-based reconstruction requires
\emph{exact} residues for multiple moduli simultaneously, and errors are frequent enough
at MMA$\approx$50\% to cause failure (Top-1 = 0.09\% in CRT mode).
Approximately 13\% of IntSeqBERT Solver calls return no valid candidate (\texttt{none} mode),
primarily due to CRT failures; approximate CRT or relaxed residue constraints are a natural remedy.

\textbf{Dataset bias.} OEIS sequences are predominantly non-negative (\texttt{nonn}).
The high sign accuracy (98.54\%) may partially reflect this bias.
Similarly, sequences with linear or sub-linear growth remain the majority even after our
filtering, and the high overall magnitude accuracy partially reflects this growth bias;
the contribution of our representation shows up within this bias as the large accuracy gains
in the Large and higher magnitude buckets (Tables~\ref{tab:scale} and~\ref{tab:solver-bucket})
and the effectiveness of the modulo spectrum (Section~\ref{sec:modulo-spectrum}).
Astronomical-bucket metrics are based on few samples and should be interpreted with caution.

\textbf{Compute constraints.} All experiments were conducted on a single consumer GPU
(GeForce RTX 3070 Ti, 8\,GB VRAM), which constrains model size and vocabulary capacity.
Due to this budget, all reported runs use a single random seed (42), and we do not provide
multi-seed variance estimates.
The Vanilla baseline's vocabulary (20,003) is smaller than that of
FACT~\cite{zurich-fact}, which handles values up to several million using larger compute
resources. Consequently, our Vanilla results represent a FACT-equivalent architecture under
our resource constraints rather than a direct reproduction of FACT's reported numbers.


\section{Conclusion}
\label{sec:conclusion}

We have presented \textbf{IntSeqBERT}, a dual-stream Transformer encoder that fuses a
continuous log-scale magnitude embedding with sin/cos modulo embeddings via FiLM, jointly
trained on 219,765 OEIS sequences. Experiments (Section~\ref{sec:experiments}) demonstrate
that the dual-stream design substantially outperforms both a standard tokenised Transformer
and a magnitude-only ablation across all scales and metrics.

Our results also carry a broader message. Even with number-theoretic structure built into
the representation, the absolute success rate of next-term prediction remains modest
(Top-1: 19.09\%), and a purely tokenised Transformer is far weaker still (2.59\%)---
confirming that neural networks struggle with direct prediction of OEIS terms.
IntSeqBERT itself is a neural-symbolic design: the network predicts magnitude, sign, and
residues separately, and the symbolic CRT solver supplies the exact number-theoretic
constraints that neural networks struggle to learn end-to-end. We therefore read our
results both as evidence of the limits of pure end-to-end learning on this task and as
motivation for hybrid symbolic--neural approaches to computer mathematics.

\textbf{Future work} includes combining magnitude uncertainty estimation with approximate CRT
to improve large-integer prediction; extending the modulo stream beyond $m = 101$;
evaluating on the remaining FACT benchmark tasks---sequence classification, sequence
similarity, next sequence-part prediction, and unmasking~\cite{zurich-fact}---to enable direct
comparison against that benchmark;
constructing family-aware dataset splits and introducing sequence-synthesis data
augmentation to improve generalisation in sparse magnitude buckets;
scaling to larger models pre-trained on the full OEIS corpus;
and exploring downstream applications such as conjecture generation,
where pretrained IntSeqBERT embeddings could serve as a rich feature backbone.


\begin{credits}
\subsubsection{\ackname}
This work was supported by JSPS KAKENHI Grant Number JP24K14897.

\end{credits}


\bibliographystyle{splncs04}
\bibliography{references}

\end{document}